\documentclass{article}

\usepackage{microtype}
\usepackage{graphicx}
\usepackage{subcaption}
\usepackage{booktabs} 

\usepackage{hyperref}

\usepackage[preprint]{icml2026}

\usepackage{amsmath}
\usepackage{amssymb}
\usepackage{mathtools}
\usepackage{amsthm}
\usepackage{enumitem}

\usepackage[capitalize,noabbrev]{cleveref}

\theoremstyle{plain}

\theoremstyle{definition}

\theoremstyle{remark}

\usepackage{booktabs}
\usepackage{multirow}
\usepackage{colortbl}
\usepackage{xcolor}

\usepackage[textsize=tiny]{todonotes}

\begin{document}

\twocolumn[
  \icmltitle{A Step Toward Federated Pretraining of Multimodal Large Language Models}



  \icmlsetsymbol{equal}{*}

  \begin{icmlauthorlist}
    \icmlauthor{Baochen Xiong}{yyy,comp,sch,equal}
    \icmlauthor{Yifan Xu}{sch2,equal}
    \icmlauthor{Xiaoshan Yang}{yyy,comp,sch}
    \icmlauthor{Yaguang Song}{comp}
    \icmlauthor{Yaowei Wang}{comp}
    \icmlauthor{Changsheng Xu}{yyy,comp,sch}
  \end{icmlauthorlist}
  \icmlaffiliation{yyy}{MAIS, Institute of Automation, Chinese Academy of Sciences}
  \icmlaffiliation{sch2}{King Abdullah University of Science and Technology (KAUST)}
  \icmlaffiliation{comp}{Pengcheng Laboratory}
  \icmlaffiliation{sch}{School of Artificial Intelligence, University of Chinese Academy of Sciences (UCAS)}

  \icmlcorrespondingauthor{Xiaoshan Yang}{xiaoshan.yang@nlpr.ia.ac.cn}


  \vskip 0.3in
]



\printAffiliationsAndNotice{}  

\begin{abstract}
The rapid evolution of Multimodal Large Language Models (MLLMs) is bottlenecked by the saturation of high-quality public data, while vast amounts of diverse multimodal data remain inaccessible in privacy-sensitive silos.
Federated Learning (FL) offers a promising solution to unlock these distributed resources, but existing research focuses predominantly on fine-tuning, leaving the foundational pre-training phase largely unexplored.
In this paper, we formally introduce the \textbf{Federated MLLM Alignment (Fed-MA)} task, a lightweight pre-training paradigm that freezes the vision encoder and LLM while collaboratively training the cross-modal projector.
We identify two critical challenges in this setting: (i) parameter interference in aggregating local projectors; and (ii) gradient oscillations in one-pass collaborative SGD.
To address these challenges, we propose \textbf{Fed-CMP}, a pioneering framework for federated MLLM pre-training.
Fed-CMP employs \textbf{Canonical Reliability-Aware Aggregation}, which constructs a canonical space to decompose client projectors into a shared alignment basis and client-specific coefficients, then performs reliability-weighted fusion to suppress parameter interference.
Furthermore, Fed-CMP introduces \textbf{Orthogonality-Preserved Momentum}, which applies momentum to the shared alignment basis via orthogonal projection, accumulating historical optimization directions while preserving geometric structure.
We construct four federated pre-training scenarios based on public datasets, and extensive experiments validate that Fed-CMP significantly outperforms existing baselines.
\end{abstract}

\section{Introduction}
%
%
Multimodal Large Language Models (MLLMs) have emerged as a promising paradigm for open-world multimodal intelligence, demonstrating exceptional capabilities in image understanding, visual question answering, and reasoning~\cite{wu2023multimodal,yin2024survey}.
The success of these models is largely fueled by the centralized training on massive-scale public image–text datasets~\cite{zhang2024mm}.
However, the scaling of MLLMs faces a looming bottleneck: high-quality public data is approaching saturation and may soon be insufficient to sustain scaling laws~\cite{jones2024ai}.
Meanwhile, vast amounts of diverse multimodal data reside in privacy-sensitive domains (such as personal devices and private institutions) remaining fragmented and inaccessible~\cite{hou2025private}.
Leveraging these distributed ``data silos'' is essential for the next stage of MLLM evolution, yet centralized collection is prohibited by strict privacy regulations and data sovereignty laws.

%

To address this data dilemma, Federated Learning (FL) offers a compelling solution.
By enabling collaborative training across decentralized clients while keeping raw data local, FL preserves privacy while leveraging distributed knowledge~\cite{mcmahan2017communication}. 
Integrating FL with MLLM pre-training has the potential to break the boundaries of public datasets, allowing models to learn from diverse, real-world distributions that are currently unseen.
However, directly applying FL to MLLMs presents significant challenges.
The sheer size of MLLMs (billions of parameters) imposes prohibitive communication and computation overheads on standard federated frameworks, making full-parameter training on edge clients challenging.
Existing research predominantly concentrates on the fine-tuning stage, leaving the foundational pre-training phase largely unexplored~\cite{xu2024fedmllm,xiong2025pilot,xuyou}.
This gap is significant because MLLMs are still in a developmental stage where scaling up training data is essential for acquiring comprehensive multimodal understanding and generalization capabilities.


In this paper, we present an initial step towards a federated pre-training framework for Multimodal Large Language Models (MLLMs).
Given the extreme time consumption associated with end-to-end pre-training of entire MLLMs, we adopt the widely studied “Visual-Projector-Language” architecture~\cite{liu2023visual,wang2024qwen2} to instantiate the Federated Pre-training problem and conduct preliminary benchmarks.
We formally introduce the \textbf{Federated MLLM Alignment} (Fed-MA) task to address the unique challenges and constraints inherent in the pre-training stage within a federated learning (FL) context.
In this framework, we freeze both the visual encoder and the language model (LLM), allowing clients to exclusively train the lightweight cross-modal projector using local image-text pairs.
This framework aligns visual features with the language space, adhering to standard MLLM pre-training paradigms while substantially reducing communication and computational overheads.
Consequently, the server only synchronizes the projector parameters, enabling efficient and privacy-preserving collaborative learning.

\begin{figure}[t]
\centering
    \includegraphics[width=1\linewidth]{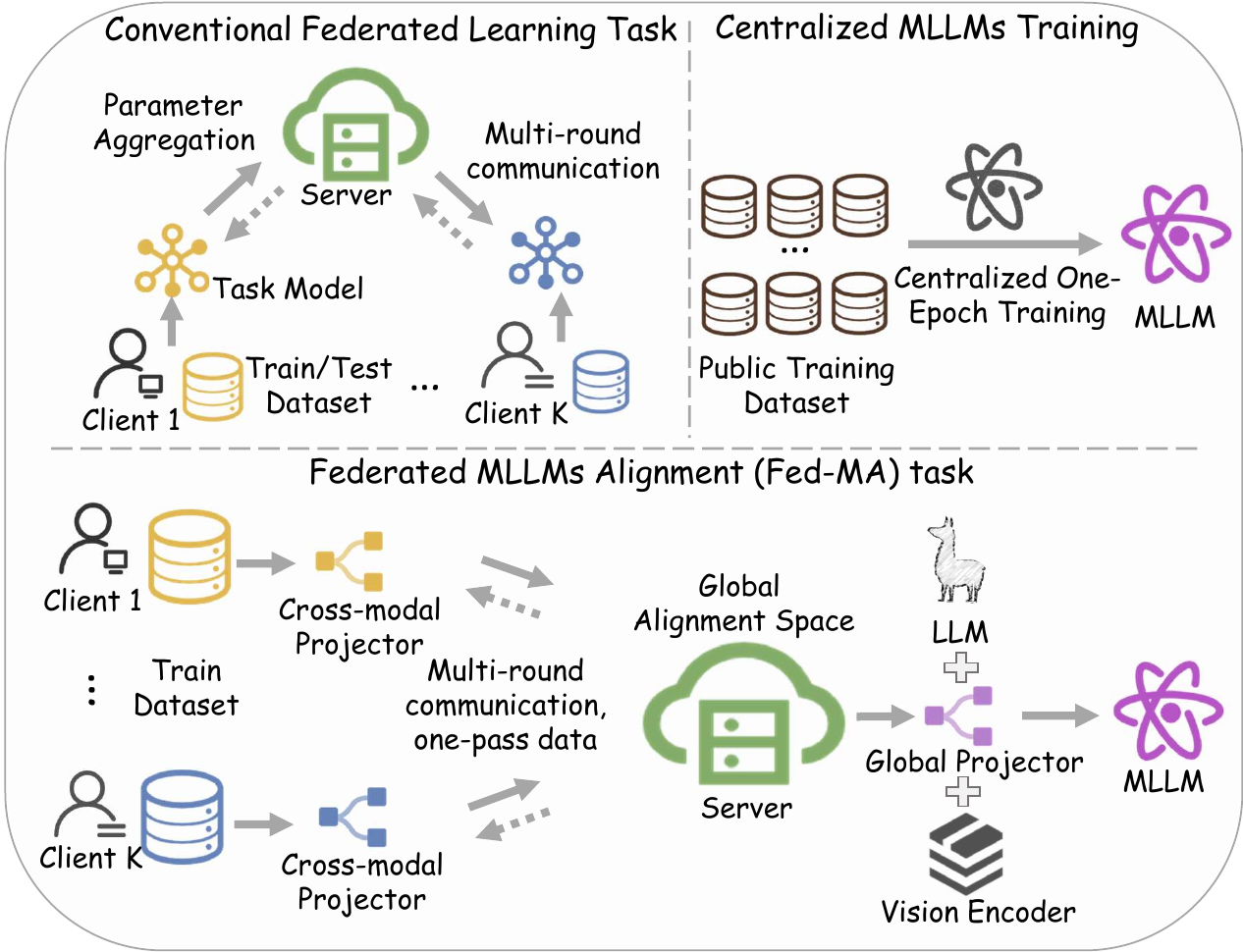} 
    \caption{A step toward Federated MLLMs: Comparison among conventional Federated Learning, centralized MLLMs training, and the proposed Federated MLLMs Alignment (Fed-MA) task.
    } 
    \vspace{-1.0em}
    \label{motivation}
\end{figure}

As illustrated in Figure~\ref{motivation}, the proposed Fed-MA task targets the cross-modal alignment stage of MLLMs, where the goal is to collaboratively train a shared cross-modal projector that maps visual features into the textual embedding space of the LLM.
Specifically, the Fed-MA task presents at least two critical challenges:
\textbf{(i) Parameter Interference in Aggregating Local Projectors.}
The Non-IID nature of multimodal data causes severe inconsistencies in the cross-modal mappings learned by different clients. 
Since clients optimize their projectors on distinct local distributions, the resulting projection directions diverge significantly—visual features are mapped to inconsistent regions of the LLM's embedding space across clients.
Directly aggregating these divergent parameters leads to destructive interference, causing the global projector to produce incoherent cross-modal mappings.
\textbf{(ii) Gradient Oscillations in One-Pass Manner Collaborative SGD.} 
In the pre-training paradigm, data is typically consumed in a one-pass manner without repetition.
Under federated settings, this means clients train on streaming, non-repeating data partitions across communication rounds, and previously seen samples are never revisited.
Consequently, the local updates at each round only reflect the gradient information of the current data, lacking memory of historical optimization directions.
When aggregated, the global projector is susceptible to being dominated by the transient gradients of the current round, leading to optimization oscillations and catastrophic forgetting of previously learned cross-modal mappings.

To take a pioneering step toward effective federated MLLM pre-training, we propose \textbf{Fed-CMP}, 
a framework designed to collaboratively align cross-modal projectors across distributed clients.
To address parameter interference, Fed-CMP employs \textbf{Canonical Reliability-Aware Aggregation (CRA)}.
Unlike standard parameter averaging that suffers from destructive interference due to divergent alignment directions, CRA constructs a canonical space where client projectors are decomposed into a shared alignment basis and client-specific coefficients.
The shared alignment basis captures the common cross-modal mapping direction, transforming aggregation from fusing conflicting parameter matrices to combining compatible coefficient vectors.
Since the reliability of client-specific coefficients varies due to heterogeneous data quality, CRA performs reliability-weighted fusion based on update magnitude and cross-modal alignment quality, suppressing parameter interference from poorly aligned clients.
To tackle gradient oscillations, we introduce \textbf{Orthogonality-Preserved Momentum (OPM)}.
We observe that the shared alignment basis captures the consensus cross-modal mapping direction and is substantially less sensitive to round-specific data perturbations.
OPM therefore applies momentum to this shared basis to leverage its stability properties.
However, since linear combinations of orthogonal matrices violate orthogonality, OPM performs orthogonal projection to keep the basis on the orthogonal manifold while accumulating historical optimization directions.
An adaptive momentum coefficient further amplifies smoothing when drastic distribution shifts are detected, preventing catastrophic forgetting.

Our contributions are summarized as follows:
First, we pioneer the exploration of Federated MLLMs Pre-training, offering a new perspective on privacy-preserving MLLMs training.
Second, we propose the Fed-CMP framework.
Specifically, Fed-CMP incorporates two key components: Canonical Reliability-Aware Aggregation (CRA), which suppresses parameter interference from heterogeneous clients, and Orthogonality-Preserved Momentum (OPM), which mitigates gradient oscillations in one-pass federated collaborative training.
Third, we construct four federated pre-training scenarios based on public datasets.
Extensive experiments validate that our method significantly outperforms existing baselines.

\section{Related Work}

\subsection{Multimodal Large Language Models}
Rapid developments have been witnessed in multimodal large language models (MLLMs) that enable human interaction with both words and visual content.
Some MLLMs extend LLMs~\cite{wu2023visual} as central controllers, integrating them with various functional agents, with language serving as a general interface.
Another series of studies explored direct training of MLLMs, including initial training using a unified architecture~\cite{comanici2025gemini}, integrating pre-trained visual encoders with LLMs through simple projection~\cite{peng2023kosmos,dai2023instructblip,liu2023visual,zhu2023minigpt} or cross-attention mechanisms~\cite{2024Libra}.
The ``Visual Encoder–Projector–LLM'' framework has witnessed rapid development in the open-source community.
During pre-training, the visual encoder and the LLM are frozen, and only the projector is optimized for multimodal alignment.
The pre-trained projector establishes a stable multimodal alignment that serves as the foundation for subsequent fine-tuning of MLLMs.
Despite these advances, existing MLLM pre-training is predominantly conducted in a centralized manner, without considering data privacy and security constraints.
Given its widespread adoption, this representative framework serves as an ideal foundation for validating our federated pre-training approach.
To this end, we explore a novel Federated Multimodal Alignment task.

\subsection{Federated Learning}
Federated learning has emerged as a promising paradigm for privacy-preserving collaborative training.
The earliest FL algorithm is FedAvg~\cite{mcmahan2017communication}, which builds the global model by averaging the local updates obtained by stochastic gradient descent (SGD).
However, FedAvg inevitably suffers performance degradation on non-IID data~\cite{liaosplitting}.
To deal with this problem,
SCAFFOLD~\cite{karimireddy2020scaffold} uses control variates (variance reduction) to correct for the client-drift in its local updates.
FedProx~\cite{li2020federated} uses regularization loss to prevent client models from deviating from the aggregated server model.
MOON~\cite{li2021model} uses the similarity between model representations to correct local training of individuals, performing comparative learning at the model level.
Furthermore, several works focus on improving generalization,
FedSAM~\cite{qu2022generalized} uses a sharpness-aware optimization strategy to improve generalization by penalizing sharp loss landscapes during local training, whereas FedSWA~\cite{liuimproving} employs a stochastic weight averaging mechanism to locate flat global minima by combining cyclical learning rates with weight averaging.
However, the above methods mainly align different local models through parameter, gradient, or representation constraints. 
Instead of focusing on constraining local updates, our method utilizes uploaded local models on the server to collaboratively construct a shared semantic space.

More recently, some studies have begun to explore FL in the context of MLLMs~\cite{zhang2025fednano,zhang2025mllm,yang2025survey,fangfedpha}.
For example, Pilot~\cite{xiong2025pilot} uses a two-stage adapter framework to enable collaborative, cross-task instruction-tuning of MLLMs across distributed clients.
FedMLLM~\cite{xu2024fedmllm} introduces a benchmark for federated fine-tuning of MLLMs in heterogeneous scenarios, 
while FedVLMBench~\cite{zheng2025fedvlmbench} expands the evaluation landscape with a systematic benchmark encompassing different VLM architectures and strategies.
YOCO~\cite{xuyou} uses directional supervision and sign-regularized LoRA to establish a one-shot federated learning framework for MLLMs that ensures global consistency.
Current research in this field primarily concentrates on the federated fine-tuning of MLLMs.
However, the pre-training stage is equally critical, as it establishes the foundational cross-modal alignment upon which all downstream fine-tuning depends. 
To this end, our work targets the pre-training stage, aiming to establish robust visual-language alignment using decentralized client data while strictly preserving privacy.

\begin{figure*}[t]
\centering
    \includegraphics[width=1\linewidth]{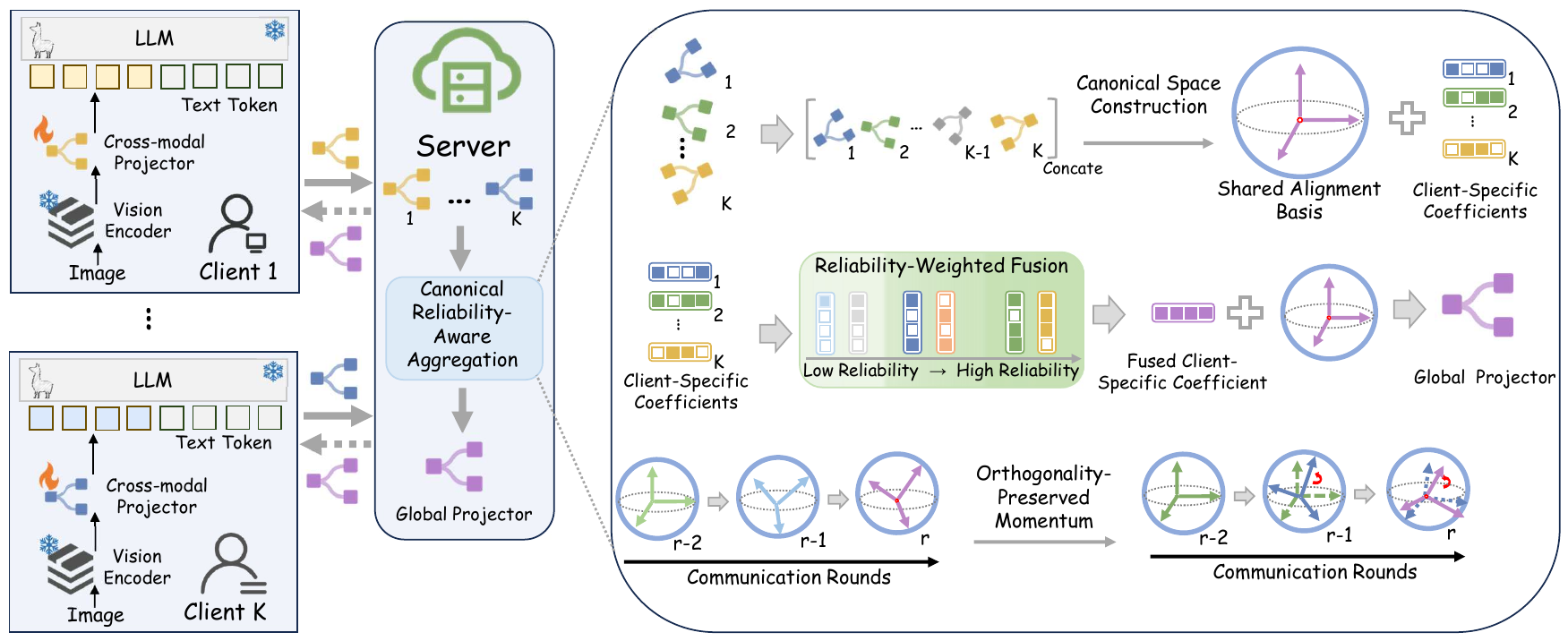}
    \caption{
    Overview of the proposed Federated MLLM pretraining framework (Fed-CMP). 
}\label{framework}
\end{figure*}

\section{Methodology}
\subsection{Task Definition}
We assume there are $K$ clients.
Each client $k$ holds a local dataset of image–text pairs $\mathcal{D}_k=\{{\bf v}_{i,k}^{img},{\bf x}_{i,k}^{t}\}_{i=1}^{n_k}$, 
where ${\bf v}_{i,k}^{img}$ denotes an image and ${\bf x}_{i,k}^{t}$ denotes its associated textual description tokens.
In the Fed-MA task, the local dataset on each client is partitioned into
$R$ disjoint subsets,
$\mathcal{D}_k=\bigcup_{r=1}^R \mathcal{D}_k^{(r)}$,
corresponding to $R$ communication rounds.
The total number of data pairs is $n=\sum_{k=1}^Kn_k$.

All clients are initialized with an identical pre-trained MLLM consisting of
a frozen visual encoder $f(\cdot)$, a frozen LLM $L$, and a trainable cross-modal projector $\psi$.
Given an input image ${\bf v}^{img}$, the visual encoder extracts visual features $H^v$.
The cross-modal projector is used to align the visual feature with the LLM.
The projector transforms $H^v$ into language embedding tokens ${\bf x}^{img}\in \mathbb{R}^{N \times C}$, where $N$ is the number of projected visual tokens and $C$ is the hidden size,
\begin{equation}
{\bf x}^{img} = \psi(H^v) \text {, with } H^v=f({\bf v}^{img}).
\end{equation}
Finally, ${\bf x}^{img}$ and ${\bf x}^{t}$ are input into the LLM to generate a response.
The Fed-MA task aims to collaboratively train a global cross-modal projector for MLLMs under the federated learning paradigm.
The task is defined over $R$ global communication rounds.
In each round, the server broadcasts the current global projector to all clients, and each client trains the projector locally for one epoch using its corresponding subset of the local dataset.
The updated projector parameters are then uploaded to the server for aggregation,
and the aggregated global projector is broadcast back to the clients for the next round.
Through multiple communication rounds, clients collaboratively learn a global
cross-modal projection module without centralized data sharing.
The overall optimization goal is defined as follows:
\begin{equation}
\sum_{k=1}^{K}\frac{n_k}{n}
\mathbb{E}_{({\bf v}^{img},{\bf x}^{t})\sim \mathcal{D}_k}
\sum_{j=1}^{L}
-\log p_\theta\left(
{\bf x}^{t}_{j} \mid {\bf x}^{img}, {\bf x}^{t}_{<j}
\right)
\end{equation}
where $L$ represents the length of the textual sequence, $\theta$ denotes the parameters of the trainable cross-modal projector.
${\bf x}^{t}_{j}$ denotes the $j$-th textual token, and
${\bf x}^{t}_{<j}$ denotes all textual tokens preceding the index $j$.



\subsection{Federated MLLM Pre-training Framework}
As shown in Figure~\ref{framework}, we propose Fed-CMP, a framework to collaboratively optimize cross-modal projectors across distributed clients.
Each client performs local autoregressive training on private image–text pairs, communicating only projector parameters to the server. 
However, this process faces two critical challenges: 
(i) parameter interference in aggregating local projectors, where Non-IID data causes clients to learn divergent cross-modal mappings, and directly aggregating these divergent projectors leads to destructive interference;
and (ii) gradient oscillations in One-Pass manner collaborative SGD, where one-pass, non-repeating data consumption causes local updates to lack memory of historical optimization directions, leading to optimization oscillations and catastrophic forgetting.
To tackle the above challenges,
Fed-CMP integrates two coupled components: \textbf{Canonical Reliability-Aware Aggregation} strategy and \textbf{Orthogonality-Preserved Momentum} update.
We detail these components below.

%
%


\subsubsection{Canonical Reliability-Aware Aggregation}

To address the parameter interference in aggregating local projectors, we propose \textbf{Canonical Reliability-Aware Aggregation (CRA)}, which performs aggregation in a carefully constructed canonical space that mitigates parameter interference across heterogeneous clients.

The core idea of CRA is to establish a canonical space where client-specific projectors can be compared and fused without directional conflicts.
In standard parameter averaging, projectors learned from heterogeneous data exhibit divergent alignment directions, leading to destructive interference during aggregation.
CRA addresses this by decomposing all client projectors into a shared alignment basis and client-specific coefficients, then performing reliability-weighted fusion in this unified space.

\textbf{Canonical Space Construction.}
Let $\mathbf{W}_{k} \in \mathbb{R}^{I \times O}$ denote the projector weight matrix from client $k$, where $I$ and $O$ represent the input and output dimensions.
To construct the canonical space, we first concatenate local projectors from all $K$ clients into a joint matrix $\Delta \mathbf{W} = [\mathbf{W}_1, \ldots, \mathbf{W}_K]$, and perform singular value decomposition following~\cite{stoica2024model}:
\begin{equation}\label{cra}
\mathrm{SVD}\left(\Delta \mathbf{W}\right)
= \mathbf{U} \boldsymbol{\Sigma}
\left[\mathbf{V}_{1}, \mathbf{V}_{2}, \cdots, \mathbf{V}_{K}\right]^{\top}.
\end{equation}
This decomposition yields two semantically meaningful components:
(1) The shared alignment basis $\mathbf{U}\boldsymbol{\Sigma}$, which captures the common cross-modal mapping direction and serves as the reference basis vectors for aggregation;
(2) The client-specific coefficients $\{\mathbf{V}_{k}\}_{k=1}^{K}$, which represent each client's projection as coefficients with respect to the shared basis.
By expressing all client projectors in terms of the same orthonormal basis $\mathbf{U}$, CRA eliminates directional divergence and transforms the aggregation problem from fusing conflicting parameter matrices to combining compatible coefficient vectors.

\textbf{Reliability-Weighted Fusion.}
While the canonical space enables valid comparison of client-specific coefficients, their reliability varies significantly due to heterogeneous data quality and quantity.
Naive averaging of $\{\mathbf{V}_{k}\}_{k=1}^{K}$ risks incorporating noise from poorly aligned clients.
To mitigate parameter interference during fusion, we introduce reliability-aware weighting that prioritizes high-reliability coefficients.

We evaluate the reliability of each client based on two complementary metrics: the update magnitude (reflecting information gain) and the cross-modal alignment quality (reflecting mapping validity).
The aggregation weight $w_k$ for client $k$ is computed as:
\begin{equation}
w_k =
\frac{
\left\| \theta_k - \theta_{0} \right\|^2 \cdot e^{-\alpha \, \mathrm{d}_k}
}{
\sum_{i=1}^{K}(
\left\| \theta_i - \theta_{0} \right\|^2 \cdot e^{-\alpha \, \mathrm{d}_i})
},
\end{equation}
where $\theta_k$ and $\theta_0$ denote the updated projector parameters and the global projector parameters, respectively, and $\alpha \geq 0$ controls the influence of alignment quality.
The term $\left\| \theta_k - \theta_{0} \right\|^2$ quantifies the update magnitude~\cite{zhou2024metagpt}, while $\mathrm{d}_k$ measures the local cross-modal alignment error:
\begin{equation}
\mathrm{d}_k
=
\frac{1}{|\mathcal{D}_k|}
\sum_{i \in \mathcal{D}_k}
\left\|
\mathbf{\tilde{x}}^{img}_{i,k}
-
\mathbf{\tilde{x}}^{t}_{i,k}
\right\|_2,
\end{equation}
where $\mathbf{\tilde{x}}^{img}_{i,k}$ and $\mathbf{\tilde{x}}^{t}_{i,k}$ denote the averaged projected visual and textual token representations of the $i$-th sample from client $k$, respectively.

Finally, the global projector is reconstructed by applying the reliability weights to the client-specific coefficients and combining with the shared alignment basis:
$\mathbf{W}_{\mathrm{global}} = \mathbf{U}\boldsymbol{\Sigma} \left[\sum_{k=1}^{K} w_k \mathbf{V}_k\right]^{\top}$.
By performing aggregation in the canonical space with reliability-aware weighting, CRA effectively suppresses parameter interference from heterogeneous and poorly aligned clients, ensuring the global projector converges towards consistent and high-quality cross-modal alignment.

\subsubsection{Orthogonality-Preserved Momentum}
To counteract the gradient oscillations, we propose \textbf{Orthogonality-Preserved Momentum (OPM)}, which applies momentum to the shared alignment basis while preserving its orthogonal structure.

In one-pass federated pre-training, each data sample is seen only once, leading to high variance in parameter updates and temporal instability across communication rounds.
Standard momentum can smooth out such oscillations and improve stability, but directly applying it to projector parameters is suboptimal due to the non-repeating data streams.
We observe that the shared alignment basis obtained from the canonical decomposition (Eq.~\ref{cra}) captures the consensus cross-modal mapping direction across clients, making it substantially less sensitive to round-specific data perturbations.
Therefore, OPM applies momentum to this shared basis to leverage its stability properties.

However, directly applying standard momentum to the orthogonal basis $\mathbf{U}$ violates its orthogonality constraint, since linear combinations of orthogonal matrices are generally non-orthogonal.
Such violations distort the geometric structure of the shared alignment basis and may lead to progressive degradation of cross-modal alignment quality.
To address this, OPM performs a two-step update consisting of alignment and orthogonal projection.
Specifically, let $\mathbf{U}_r$ denote the global orthogonal basis at round $r$, and $\hat{\mathbf{U}}_{r+1}$ denote the newly aggregated basis from the current round.
We update $\mathbf{U}_{r+1}$ as:
\begin{equation}
\mathbf{U}_{r+1} 
= \mathrm{Orth}\!\big(
\beta \mathbf{U}_r + (1-\beta)\, \mathrm{Align}(\hat{\mathbf{U}}_{r+1}, \mathbf{U}_r) \big),
\end{equation}
where $\beta \in [0,1)$ is the momentum coefficient.
The operator $\mathrm{Align}(\cdot)$ resolves the inherent sign ambiguity of singular vectors by aligning each column of $\hat{\mathbf{U}}_{r+1}$ with the corresponding column of $\mathbf{U}_r$, flipping the sign whenever their inner product is negative.
The operator $\mathrm{Orth}(\cdot)$ projects the resulting matrix onto the orthogonal manifold via polar decomposition.
We denote the momentum-mixed matrix by $\mathbf{M}=\beta \mathbf{U}_r + (1-\beta)\, \mathrm{Align}(\hat{\mathbf{U}}_{r+1}, \mathbf{U}_r)$.
Concretely, the update of the orthogonal basis $\mathbf{U}_{r+1}$ is given by:
\begin{equation}
\mathbf{U}_{r+1}
=
\mathrm{Orth}(\mathbf{M})
=
\mathbf{P}\mathbf{Q}^\top,
\quad
\text{where }
\mathbf{M} = \mathbf{P}\boldsymbol{\Lambda}\mathbf{Q}^\top.
\end{equation}
The projection yields the closest orthogonal matrix to $\mathbf{M}$ and guarantees that $\mathbf{U}_{r+1}$ remains on the orthogonal manifold.
We provide the detailed derivation in Appendix~\ref{appendix:dcm_proof}.

For the singular values $\boldsymbol{\Sigma}$, which are unconstrained scalar coefficients and do not require orthogonality preservation, we apply a standard momentum update.

To further enhance stability under Non-IID data distributions, we adopt an adaptive momentum coefficient.
Specifically, the momentum factor $\beta$ is dynamically adjusted based on the cosine dissimilarity between $\hat{\mathbf{U}}_{r+1}$ and $\mathbf{U}_r$:
\begin{equation}
\beta_{\text{adaptive}}
=
\beta_{\min}
+
(\beta_{\max}-\beta_{\min})
\left(1 - \mathrm{CosDis}^{\lambda}\right),
\end{equation}
where $\mathrm{CosDis}$ measures the average cosine dissimilarity between corresponding columns, $\lambda$ controls sensitivity, and $\beta_{\min}$ and $\beta_{\max}$ bound the momentum range.

Through these operations, OPM accumulates the historical evolution of the consensus direction, dampening oscillations and improving the stability of global cross-modal alignment.

\section{Experiments}
\subsection{Experimental Setups}
\begin{figure}[t]
\centering
    \includegraphics[width=\linewidth]{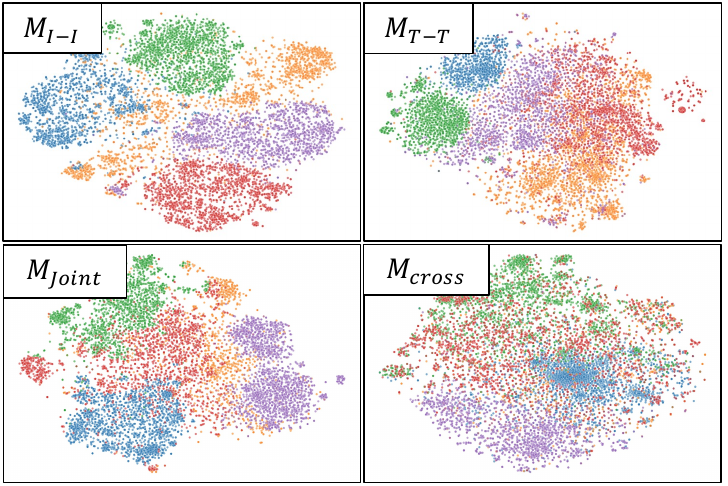}
    \caption{
    t-SNE visualization of heterogeneous clustering on the CC12M dataset, utilizing 5 distinct clusters across 4 similarity metrics. 
}\label{fig:tsne_concise}
\vspace{-1.3em}
\end{figure}

To verify the effectiveness of the proposed Fed‑CMP method on the Federated MLLMs Alignment (Fed‑MA) task, 
we construct \textbf{four} federated multimodal pretraining datasets based on CC12M~\cite{cc12m}.
Specifically, we partition CC12M into $5$ clients using a heterogeneous clustering strategy, resulting in \textbf{four} representative clustering scenarios: Image–Image Clustering ($\mathcal{M}_{I-I}$), Text–Text Clustering ($\mathcal{M}_{T-T}$), Joint Image-Text Clustering ($\mathcal{M}_{joint}$), and Cross‑Modal Image–Text Clustering ($\mathcal{M}_{cross}$).  
These settings simulate different types of modality‑induced statistical heterogeneity in federated multimodal pretraining.
To illustrate the data distribution characteristics, we visualize the client data using t-SNE~\cite{tsne} in Figure~\ref{fig:tsne_concise}, where each color represents one of the five clients.
From the visualization, we observe that the degree of heterogeneity follows the order: $\mathcal{M}_{T-T} < \mathcal{M}_{cross} < \mathcal{M}_{I-I} \approx \mathcal{M}_{joint}$, indicating that text-text clustering yields relatively homogeneous partitions, while image-based clustering introduces more severe distribution shifts across clients.
Dataset construction procedures are provided in Appendix~\ref{dataset}.

\textbf{Models and Benchmark Datasets.} 
In all experiments, we follow the LLaVA~\cite{liu2023visual} architecture for the multimodal large language model.
Specifically, we adopt a pretrained CLIP ViT-L/14~\cite{clip} as the visual encoder, Vicuna-v1.5-7B~\cite{chiang2023vicuna} as the language model, and a two-layer MLP as the cross-modal projector.
Following the pre-training paradigm, the visual encoder and language model are kept frozen, and only the projector is trained through federated collaboration.
We evaluate all methods on seven multimodal benchmarks, including MM-Vet, MMBench, SEED, LLaVA-Bench, POPE, MME, and MMVP.

\textbf{Baselines.}
We compare our method with eight representative baselines, including
non‑federated training, widely used federated learning algorithms,
and model merging algorithms,
detailed as follows:
\begin{itemize}[leftmargin=1.2em, itemsep=0pt, topsep=2pt]
    \item \textbf{Non‑Federated Training.}
    Each client trains its model independently, without any federated aggregation.
    \item \textbf{FL Algorithms.}
    We categorize the considered FL baselines into two groups:
    \textbf{Adaptive Optimization FL}, including \textbf{FedAdam}~\cite{reddi2020adaptive},
    which enhances aggregation with adaptive or momentum;
    and \textbf{Weighted Averaging-Based FL}, comprising
    \textbf{FedAvg}~\cite{mcmahan2017communication},
    \textbf{FedProx}~\cite{li2020federated},
    and \textbf{MOON}~\cite{li2021model},
    which focus on parameter averaging, regularization, and contrastive model alignment.
    \item \textbf{Model Merging Algorithms.}
    In addition, we combine FedAvg with three recent model merging techniques, Task Arithmetic (TA)~\cite{ilharco2022editing}, TIEs~\cite{yadav2023ties}, and DARE~\cite{yu2024language}, to evaluate their effectiveness under the Fed-MA setting.
\end{itemize}

\definecolor{oursblue}{RGB}{220,230,250}
\definecolor{lightgray}{RGB}{245,245,245}

\begin{table*}[t]
\centering
\small
\setlength{\tabcolsep}{10pt}
\caption{Evaluation of MLLMs on multimodal benchmarks under the $\mathcal{M}_{I-I}$ setting.}\label{table1}
\begin{tabular}{lccccccc}
\toprule
Method & MM-Vet & MMBench & SEED & LLaVABench & POPE & MME & MMVP \\
\midrule
Local Training
& 26.9 & 25.8 & 24.5 & 35.9 & 60.9 & 856.4 & 29.9 \\

FedAvg
& 20.5 & 23.7 & 26.6 & 40.0 & 74.3 & 1056.4 & \underline{33.6} \\

FedAdam
& 19.4 & 24.6 & 25.3 & 39.2 & 69.6 & 1002.5 & 32.4 \\

FedProx
& 20.0 & 21.4 & 21.9 & 38.4 & 68.4 & 991.6 & 30.3 \\

MOON
& 22.5 & 25.5 & 26.9 & 41.2 & 73.5 & 937.6 & 31.9 \\
\rowcolor{lightgray}
FedAvg-TA
& 23.6 & 24.7 & 24.6 & 40.7 & 68.6 & 915.5 & \textbf{33.9} \\
\rowcolor{lightgray}
FedAvg-TIEs
& \underline{28.5} & \underline{29.6} & \underline{27.2} & 41.3 & \textbf{75.0} & \underline{1095.6} & 31.4 \\
\rowcolor{lightgray}
FedAvg-DARE
& 26.7 & 23.5 & 22.8 & \underline{42.5} & 69.5 & 806.5 & 26.5 \\

\rowcolor{oursblue}
Fed-CMP
& \textbf{32.5} & \textbf{32.7} & \textbf{30.5} & \textbf{46.6} & \underline{74.5} & \textbf{1143.2} & 33.0 \\
\bottomrule
\end{tabular}
\end{table*}

\begin{table*}[t]
\centering
\small
\setlength{\tabcolsep}{10pt}
\caption{Evaluation of MLLMs on multimodal benchmarks under the $\mathcal{M}_{T-T}$ setting.}\label{table2}
\begin{tabular}{lccccccc}
\toprule
Method & MM-Vet & MMBench & SEED & LLaVABench & POPE & MME & MMVP \\
\midrule
Local Training
& 25.4 & 30.1 & 19.6 & 38.5 & 72.8 & 958.9 & 30.5 \\

FedAvg
& 25.8 & 33.9 & 20.3 & 39.6 & \underline{74.9} & \underline{1059.4} & 35.2 \\

FedAdam
& 24.6 & \underline{34.6} & 20.2 & 39.2 & 72.5 & 968.1 & 33.9 \\

FedProx
& 22.9 & 29.1 & 18.4 & 36.9 & 70.8 & 871.6 & 28.2 \\

MOON
& 25.6 & 34.5 & 22.7 & 40.1 & 72.0 & 989.3 & 36.4 \\
\rowcolor{lightgray}
FedAvg-TA
& 26.3 & 31.0 & 21.2 & 40.0 & 73.1 & 982.6 & 36.6 \\
\rowcolor{lightgray}
FedAvg-TIEs
& \underline{27.6} & 32.8 & \bf{24.0} & \underline{41.5} & 74.6 & 975.8 & \textbf{37.3} \\
\rowcolor{lightgray}
FedAvg-DARE
& 25.1 & 29.6 & 15.7 & 40.6 & 66.0 & 562.4 & 36.0 \\

\rowcolor{oursblue}
Fed-CMP
& \textbf{29.9} & \textbf{35.2} & \underline{23.6} & \textbf{43.2} & \textbf{76.2} & \textbf{1075.7} & \underline{36.9} \\
\bottomrule
\end{tabular}
\end{table*}

\textbf{Training Setup.}
In all experiments, the visual encoder and the large language model are initialized from pretrained weights, while the cross‑modal projector is
randomly initialized.
We consider a federated setting with $K=5$ clients and $R=10$ communication rounds.
For each client, the local dataset is divided into 10 parts, and in each round of communication, the client will perform training on one of the data subsets for one epoch.
We train the projector with a learning rate of $1\times10^{-3}$, using a cosine learning rate scheduler and a warmup ratio of 0.03.
For the aggregation weighting, we set $\alpha=1$ by default to balance the effects of update magnitude and cross‑modal alignment quality.
For the adaptive momentum update, we set $\beta_{\min}=0.5$ and $\beta_{\max}=0.9$, with $\lambda=1$ controlling the sensitivity to cosine
dissimilarity.
All experiments were conducted on two machines, each equipped with 8 NVIDIA A100 (40 GB) GPUs, and each experiment took approximately 10 hours.

\subsection{Experimental Results}
We conduct experiments under four different clustering settings.
The main paper reports the results for $\mathcal{M}_{I-I}$ and $\mathcal{M}_{T-T}$ settings,
other experimental results are in the Appendix~\ref{additional experiment}.
For local training, results are averaged over five individual clients,
while all other results correspond to the final global model.
The best and second-best results are highlighted in bold and underlined, respectively.

As shown in Tables~\ref{table1} and~\ref{table2}, Fed-CMP achieves the best performance on most multimodal benchmarks.
Under the $\mathcal{M}_{I-I}$ setting, Fed-CMP achieves better performance,
where client data distributions are highly heterogeneous.
Under the $\mathcal{M}_{T-T}$ setting, performance differences among federated methods become smaller
due to reduced data heterogeneity.
These results indicate that Fed-CMP effectively suppresses parameter interference and mitigates gradient oscillations, thereby enhancing model generalization and cross-modal alignment across heterogeneous clients.
Fed-CMP consistently outperforms local training across all settings,
demonstrating its ability to aggregate knowledge from multiple clients
and learn models with better performance.
In addition, FedAvg-TIEs achieves the second-best performance across multiple tasks,
suggesting that incorporating model merging into federated learning
is an effective strategy for tackling the Fed-MA problem.
FedProx shows limited performance on several benchmarks, indicating that local regularization may not be well suited for Fed-MA task.

\begin{table}[t]
\centering
\caption{Ablation studies under the $\mathcal{M}_{I-I}$ setting.}\label{table3}
\small
\setlength{\tabcolsep}{4pt}
\begin{tabular}{cccccc}
\toprule
\multicolumn{3}{c}{Ablation Modules} & \multicolumn{3}{c}{Evaluation Benchmarks} \\
\cmidrule(lr){1-3} \cmidrule(lr){4-6}
CSC & RWF & OPM & LLaVA-Bench & MMVet & MMBench \\
\midrule
$\times$ & $\times$ & $\times$ & 40.0 & 20.5 & 23.7 \\
$\checkmark$ & $\times$ & $\times$ & 43.1 & 26.2 & 28.2 \\
$\times$ & $\checkmark$ & $\times$ & 43.2 & 28.7 & 29.0 \\
$\times$ & $\times$ & $\checkmark$ & 40.2 & 21.3 & 25.0 \\
$\checkmark$ & $\checkmark$ & $\times$ & 45.3 & 30.8 & 31.6 \\
$\checkmark$ & $\times$ & $\checkmark$ & 44.9 & 30.4 & 31.0 \\
\rowcolor{oursblue}
$\checkmark$ & $\checkmark$ & $\checkmark$ & \textbf{46.6} & \textbf{32.5} & \textbf{32.7} \\
\bottomrule
\end{tabular}
\end{table}

\begin{figure*}[t]
\centering
    \includegraphics[width=1\linewidth]{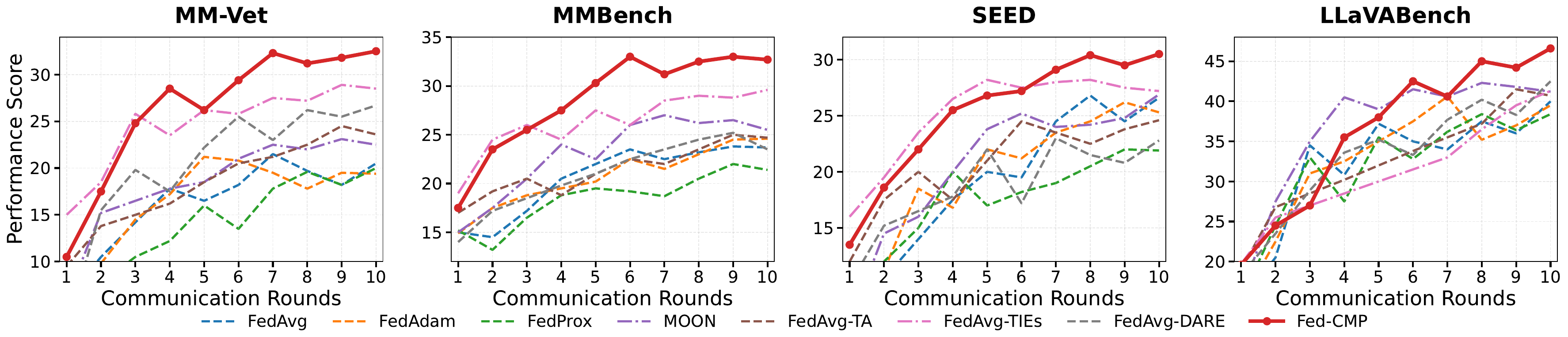}
    \caption{Performance scores across communication rounds under the $\mathcal{M}_{I-I}$ setting.
}\label{figure4}
\vspace{-1em}
\end{figure*}

\subsection{Ablation Studies}
We report ablation results under the $\mathcal{M}_{I-I}$ setting to evaluate the effectiveness of the main components of our framework.
Specifically, we study the impact of Canonical Space Construction (CSC), Reliability-Weighted Fusion (RWF), and Orthogonality-Preserved Momentum (OPM).
When RWF is removed, we replace it with standard weighted averaging.

The results are shown in Table~\ref{table3}.
Without CSC, RWF, and OPM, performance on MM-Vet and MMBench drops below local training, indicating that the model fails to handle parameter interference and gradient oscillations.
When RWF or OPM is applied alone to the raw projector parameters without CSC, performance improves in both cases, confirming the presence of the challenges these components target in the Fed-MA setting.
After introducing CSC, performance further improves across benchmarks.
By decoupling model parameters into a shared alignment basis and client-specific coefficients, CSC provides an effective foundation for suppressing parameter interference.
Moreover, RWF and OPM show complementary gains when combined with CSC, demonstrating that these components work synergistically.
However, when CSC is removed, OPM shows limited improvement, as momentum is applied directly to raw parameters without the decomposed shared basis.
This indicates that OPM is most effective when operating on the shared basis obtained by CSC, where it helps maintain update stability.
We also observe that RWF alone brings notable performance gains, while its combination with CSC yields smaller additional improvements.
We attribute this to CSC capturing most parameter discrepancies in the client-specific coefficients, which reduces the overall impact of RWF.

\begin{table}[h]
    \centering
    \small 
    \caption{Experimental analysis of RWF components and sensitivity analysis of $\alpha$ under the $\mathcal{M}_{I-I}$ setting.}
    \label{table4}
    \setlength{\tabcolsep}{10pt} 
    \begin{tabular}{cccccc}
        \toprule
        \multicolumn{3}{c}{Experimental Settings} & \multicolumn{2}{c}{Evaluation Benchmarks} \\
        \cmidrule(r){1-3} \cmidrule(l){4-5}
        Mag & Qua & \textbf{$\alpha$} & MM-Vet & MMBench \\
        \midrule
        $\times$ & $\times$ & $-$ & 30.4 & 31.0\\
        $\checkmark$ & $\times$ & $-$ & 31.2 & 31.5 \\
        $\times$ & $\checkmark$ & $-$ & 31.6 & 31.7 \\
        \midrule
        $\checkmark$ & $\checkmark$ & 0.5 & 32.0 & 32.1 \\
        \rowcolor{oursblue}
        $\checkmark$ & $\checkmark$ & 1.0 & \textbf{32.5} & \textbf{32.7} \\
        $\checkmark$ & $\checkmark$ & 2.0 & 31.6 & 32.0 \\
        \bottomrule
    \vspace{-1.2em}
    \end{tabular}
\end{table}

\section{Discussion}

\subsection{Analysis of Reliability-Weighted Fusion}
To better understand the RWF module, we analyze two key aspects: the individual contributions of its core components (update magnitude and alignment quality) and the sensitivity to the hyperparameter $\alpha$.
The experimental results are shown in Table~\ref{table4}.

\textbf{Effectiveness of Reliability Components.}
We first evaluate the performance of RWF when relying solely on update magnitude (RWF-Mag) or alignment quality (RWF-Qua). 
%
%
Compared to the baseline Fed-CMP (No-RWF), using either metric alone yields consistent performance improvements.
This confirms that both metrics are crucial for efficient parameter aggregation in Fed-MA task.
The full RWF achieves optimal performance by integrating Mag and Qua, demonstrating their complementarity.

\textbf{Impact of Hyperparameter $\alpha$.} 
Next, we investigate the impact of the hyperparameter $\alpha$ on RWF.
Results show that increasing $\alpha$ initially leads to significant performance gains.
This highlights the importance of alignment quality constraints.
Relying solely on magnitude may introduce parameters with high intensity but deviated directions, compromising global model stability.
However, performance degrades when $\alpha$ becomes excessively large.
An excessive $\alpha$ over-penalizes parameters with low alignment scores, potentially discarding valuable client-specific information and reducing generalization.
Consequently, we set $\alpha=1$ for all subsequent experiments.

\subsection{Analysis of Temporal Stability}
Figure~\ref{figure4} presents the performance trajectory of each method across 10 communication rounds.
Most baselines exhibit substantial fluctuations, with performance oscillating unpredictably between rounds—a clear manifestation of the temporal instability caused by one-pass streaming data.
For instance, FedAvg-TIEs achieves competitive scores at round 5 on SEED but drops sharply in subsequent rounds, reflecting the lack of historical momentum to stabilize the global projector.
In contrast, Fed-CMP demonstrates a consistently ascending trajectory across all four benchmarks.
This stability stems from Orthogonality-Preserved Momentum, which accumulates the consensus alignment direction across rounds and dampens abrupt updates caused by distribution shifts.
While some baselines occasionally match or exceed our method in individual rounds, they fail to maintain such performance, ultimately converging to lower plateaus.
Fed-CMP not only achieves the highest final performance but also exhibits the smoothest optimization path, validating the effectiveness of OPM in addressing gradient oscillations .

\section{Conclusion}
In this work, we bridge the gap between Federated Learning and MLLMs pre-training by introducing Fed-CMP, a framework designed to leverage distributed, privacy-sensitive data for cross-modal alignment.
We formally define the Fed-MA task and highlight its unique challenges: parameter interference in aggregating local projectors and gradient oscillations in one-pass federated collaborative training.
To overcome these challenges, Fed-CMP incorporates Canonical Reliability-Aware Aggregation, which suppresses parameter interference from heterogeneous clients, and Orthogonality-Preserved Momentum , which mitigates gradient oscillations and prevents catastrophic forgetting.
We hope this work inspires further exploration into scaling MLLMs beyond the boundaries of centralized public datasets.


\section*{Impact Statement}
This work advances the field of machine learning by introducing a federated pre-training framework for Multimodal Large Language Models (MLLMs).
We believe this work primarily addresses two pressing societal concerns.
First, it aligns with the increasing demand for user privacy across diverse technological domains by enabling collaborative model training without centralizing sensitive multimodal data from personal devices or private institutions.
Second, it offers a pathway to overcome the data bottleneck faced by current MLLMs, allowing models to learn from vast, diverse, real-world distributions that are otherwise inaccessible due to privacy regulations and data sovereignty laws.
By unlocking the potential of distributed ``data silos" in a privacy-preserving manner, this research may democratize access to high-quality MLLM training and broaden the applicability of multimodal intelligence to domains where data sharing is prohibited. 
There are many potential societal consequences of our work, none of which we feel must be specifically highlighted here.

\nocite{langley00}
\bibliography{icmlbib}
\bibliographystyle{icml2026}

\newpage
\appendix
\onecolumn
\section{Theoretical Analysis of Orthogonality-Preserved Momentum}
\label{appendix:dcm_proof}
In this section, we provide the rigorous mathematical derivation for the \textbf{Orthogonality-Preserved Momentum (OPM)} update rule.
Specifically, we address two critical geometric challenges in shared alignment basis: (1) the sign ambiguity of singular vectors (handled by the $\mathrm{Align}$ operator), and (2) the preservation of the orthogonality constraint on the Stiefel manifold (handled by the $\mathrm{Orth}$ operator).

Recall the update rule for the orthogonal basis at round $r+1$:
\begin{equation}
    \mathbf{U}_{r+1} = \mathrm{Orth}\left( \mathbf{M} \right), \quad \text{where } \mathbf{M} = \beta \mathbf{U}_r + (1-\beta)\, \mathrm{Align}(\hat{\mathbf{U}}_{r+1}, \mathbf{U}_r).
\end{equation}

\subsection{Sign Alignment Mechanism ($\mathrm{Align}$)}
\label{subsec:align_proof}

\textbf{Problem Formulation:}
Singular Value Decomposition possesses an inherent sign indeterminacy. 
For any valid decomposition $\mathbf{A} = \mathbf{U}\boldsymbol{\Sigma}\mathbf{V}^\top$, simultaneously replacing any column $\mathbf{u}_i$ with $-\mathbf{u}_i$ and $\mathbf{v}_i$ with $-\mathbf{v}_i$ yields another valid SVD:
\begin{equation}
    \mathbf{A}\mathbf{v}_i = \sigma_i \mathbf{u}_i 
    \quad \Longrightarrow \quad 
    \mathbf{A}(-\mathbf{v}_i) = \sigma_i (-\mathbf{u}_i).
\end{equation}
Consequently, the aggregated basis $\hat{\mathbf{U}}_{r+1}$ may contain column vectors that are mathematically equivalent to $\mathbf{U}_r$ but point in opposite directions (anti-parallel). Directly averaging them would result in cancellation (destructive interference) rather than momentum accumulation.

\textbf{Mechanism:}
The $\mathrm{Align}(\hat{\mathbf{U}}_{r+1}, \mathbf{U}_r)$ operator ensures directional consistency by examining the inner product between corresponding columns. Let $\mathbf{u}_{r}^{(i)}$ and $\hat{\mathbf{u}}_{r+1}^{(i)}$ denote the $i$-th column vectors of $\mathbf{U}_r$ and $\hat{\mathbf{U}}_{r+1}$, respectively.
The alignment operation is defined as:
\begin{equation}
    \hat{\mathbf{u}}_{r+1}^{(i)} \leftarrow 
    \begin{cases} 
    -\hat{\mathbf{u}}_{r+1}^{(i)} & \text{if } \langle \mathbf{u}_{r}^{(i)}, \hat{\mathbf{u}}_{r+1}^{(i)} \rangle < 0, \\
    \hat{\mathbf{u}}_{r+1}^{(i)} & \text{otherwise}.
    \end{cases}
\end{equation}
This operation flips the sign of the incoming vector if the angle between the current and historical direction exceeds $90^\circ$, ensuring that the momentum update occurs in a consistent half-space.
As shown above, this sign flip does not alter the component spanned by the basis or the underlying physical information extracted by SVD, but it is crucial for stable linear interpolation.

\subsection{Orthogonal Projection via Polar Decomposition ($\mathrm{Orth}$)}
\label{subsec:orth_proof}
\textbf{Problem Formulation:}
The momentum-mixed matrix $\mathbf{M} = \beta \mathbf{U}_r + (1-\beta) \hat{\mathbf{U}}_{r+1}$ is a linear combination of two orthogonal matrices. In Euclidean space, such a combination generally results in a matrix that is no longer orthogonal (i.e., $\mathbf{M}^\top \mathbf{M} \neq \mathbf{I}$). Geometrically, this represents a ``shrinkage" or deformation of the basis vectors. To maintain the validity of the projection component, we must project $\mathbf{M}$ back onto the closest point on the orthogonal manifold.

\textbf{Optimization Objective:}
We seek the orthogonal matrix $\mathbf{U}_{r+1}$ (denoted as $\mathbf{X}$ for derivation purposes) that minimizes the Frobenius norm distance to $\mathbf{M}$. This is formally the \textit{Orthogonal Procrustes Problem}:
\begin{equation}
    \min_{\mathbf{X}} \|\mathbf{M} - \mathbf{X}\|_F^2 \quad \text{s.t.} \quad \mathbf{X}^\top \mathbf{X} = \mathbf{I}.
\end{equation}

\textbf{Derivation:}
First, we expand the Frobenius norm objective:
\begin{equation}
\begin{aligned}
    \|\mathbf{M} - \mathbf{X}\|_F^2 &= \mathrm{Tr}\left( (\mathbf{M} - \mathbf{X})^\top (\mathbf{M} - \mathbf{X}) \right) \\
    &= \mathrm{Tr}\left( \mathbf{M}^\top \mathbf{M} - \mathbf{M}^\top \mathbf{X} - \mathbf{X}^\top \mathbf{M} + \mathbf{X}^\top \mathbf{X} \right) \\
    &= \mathrm{Tr}(\mathbf{M}^\top \mathbf{M}) - 2\mathrm{Tr}(\mathbf{M}^\top \mathbf{X}) + \mathrm{Tr}(\mathbf{I}).
\end{aligned}
\end{equation}
Since $\mathbf{M}$ is fixed and $\mathbf{X}^\top \mathbf{X} = \mathbf{I}$, the terms $\mathrm{Tr}(\mathbf{M}^\top \mathbf{M})$ and $\mathrm{Tr}(\mathbf{I})$ are constants. Therefore, minimizing the distance is equivalent to maximizing the cross-term:
\begin{equation}
    \max_{\mathbf{X}} \mathrm{Tr}(\mathbf{M}^\top \mathbf{X}).
\end{equation}

Let the Singular Value Decomposition of the mixed matrix be $\mathbf{M} = \mathbf{P} \boldsymbol{\Lambda} \mathbf{Q}^\top$, where $\mathbf{P}$ and $\mathbf{Q}$ are orthogonal matrices and $\boldsymbol{\Lambda}$ is the diagonal matrix of singular values. Substituting this into the objective:
\begin{equation}
    \mathrm{Tr}(\mathbf{M}^\top \mathbf{X}) = \mathrm{Tr}\left( (\mathbf{P} \boldsymbol{\Lambda} \mathbf{Q}^\top)^\top \mathbf{X} \right) = \mathrm{Tr}\left( \mathbf{Q} \boldsymbol{\Lambda} \mathbf{P}^\top \mathbf{X} \right).
\end{equation}
Using the cyclic property of the trace operator, $\mathrm{Tr}(\mathbf{A}\mathbf{B}\mathbf{C}) = \mathrm{Tr}(\mathbf{B}\mathbf{C}\mathbf{A})$, we rearrange the terms:
\begin{equation}
    \mathrm{Tr}\left( \boldsymbol{\Lambda} \mathbf{P}^\top \mathbf{X} \mathbf{Q} \right).
\end{equation}
Let us define a new matrix $\mathbf{Z} = \mathbf{P}^\top \mathbf{X} \mathbf{Q}$. Since $\mathbf{P}$, $\mathbf{X}$, and $\mathbf{Q}$ are all orthogonal matrices, their product $\mathbf{Z}$ is also an orthogonal matrix. The optimization problem transforms into:
\begin{equation}
    \max_{\mathbf{Z}} \mathrm{Tr}(\boldsymbol{\Lambda} \mathbf{Z}) = \max_{\mathbf{Z}} \sum_{i} \sigma_i Z_{ii},
\end{equation}
where $\sigma_i$ are the non-negative singular values from $\boldsymbol{\Lambda}$.
Because $\mathbf{Z}$ is orthogonal, its entries are bounded such that $|Z_{ii}| \le 1$. To maximize the summation with non-negative weights $\sigma_i$, we must have $Z_{ii} = 1$ for all $i$. This implies that $\mathbf{Z}$ must be the identity matrix $\mathbf{I}$.

Finally, we solve for the optimal $\mathbf{X}$:
\begin{equation}
    \mathbf{Z} = \mathbf{I} \implies \mathbf{P}^\top \mathbf{X} \mathbf{Q} = \mathbf{I} \implies \mathbf{X} = \mathbf{P}\mathbf{Q}^\top.
\end{equation}

The closest orthogonal matrix to the momentum mixture $\mathbf{M}$ is given by $\mathbf{U}_{r+1} = \mathbf{P}\mathbf{Q}^\top$. This result is equivalent to the rotational component of the Polar Decomposition of $\mathbf{M}$.
By discarding the ``stretch" component $\boldsymbol{\Lambda}$ (which represents the deformation caused by linear averaging) and retaining only the ``rotation" component $\mathbf{P}\mathbf{Q}^\top$, the $\mathrm{Orth}$ operator successfully projects the updated basis back onto the Stiefel manifold while preserving the directional semantics accumulated via momentum.

\section{A Toy Experiment for Instruction Tuning}
\label{sec:instruction_tuning}
To further validate the effectiveness of our federated pre-training framework, we conduct a preliminary experiment on downstream instruction tuning.
Specifically, we distribute the global projector obtained from Fed-CMP pre-training to all clients, and partition instruction tuning datasets among 5 clients.
Following the standard instruction tuning paradigm, each client fine-tunes both the projector and LoRA adapters attached to the LLM, while the server performs conventional weighted aggregation~\cite{zheng2025fedvlmbench}.

\subsection{Experimental Setup.}
We evaluate on two instruction tuning benchmarks: GQA~\cite{hudson2019gqa} and OCRVQA~\cite{lu2022learn}.
Each dataset is evenly partitioned into 5 non-overlapping subsets and distributed to clients.
Table~\ref{sft_data} summarizes the data statistics for each client.
We will compare the model initialized using the Fed-CMP pre-trained projector with the model initialized using the FedAvg and FedAvg-TIEs projectors.

\subsection{Results and Analysis.}
As shown in Table~\ref{sft_results}, the model initialized using our proposed Fed-CMP projector consistently outperforms the baseline model on both datasets.
This improvement demonstrates that the well-aligned cross-modal projector learned during federated pre-training provides a stronger foundation for subsequent tasks.

\begin{table}[t]
\centering
\small
\caption{Data statistics for instruction tuning experiments.}
\label{sft_data}
\begin{tabular}{l|ccccc|c|c|c}
\toprule
\textbf{Datasets} & \textbf{Client 1} & \textbf{Client 2} & \textbf{Client 3} & \textbf{Client 4} & \textbf{Client 5} & \textbf{Total} & \textbf{Test Set} & \textbf{Metric} \\
\midrule
GQA & 185K & 198K & 176K & 192K & 169K & 0.92M & Test-balanced (12.6k) & EM $\uparrow$ \\
OCRVQA & 162K & 178K & 145K & 168K & 147K & 0.80M & Test (100k) & Accuracy $\uparrow$ \\
\bottomrule
\end{tabular}
\end{table}

\begin{table}[t]
\centering
\small
\caption{Instruction tuning performance across clients.}
\label{sft_results}
\begin{tabular}{l|ccccc|ccccc}
\toprule
\multirow{2}{*}{\textbf{Method}} & \multicolumn{5}{c|}{\textbf{GQA}} & \multicolumn{5}{c}{\textbf{OCRVQA}} \\
& C1 & C2 & C3 & C4 & C5 & C1 & C2 & C3 & C4 & C5 \\
\midrule
FedAvg & 47.8 & 52.4 & 50.9 & 50.2 & 49.6 & 46.5 & 42.1 & 39.8 & 40.6 & 45.9 \\
FedAvg-TIEs & 48.2 & 53.5 & 51.3 & 51.4 & 50.0 & 47.8 & 42.5 & 40.9 & 41.0 & 47.2 \\
\rowcolor{oursblue} Fed-CMP (Ours) & \textbf{49.6} & \textbf{54.3} & \textbf{52.1} & \textbf{52.5} & \textbf{50.8} & \textbf{48.9} & \textbf{44.2} & \textbf{41.7} & \textbf{42.8} & \textbf{48.1} \\
\bottomrule
\end{tabular}
\end{table}

\begin{table*}[t]
\centering
\small
\setlength{\tabcolsep}{10pt}
\caption{Evaluation of MLLMs on multimodal benchmarks under the $\mathcal{M}_{Joint}$ setting.}\label{table_it_it}
\begin{tabular}{lccccccc}
\toprule
Method & MM-Vet & MMBench & SEED & LLaVABench & POPE & MME & MMVP \\
\midrule
Local Training
& 27.2 & 28.9 & 23.6 & 37.3 & 66.5 & 1000.8 & 31.6 \\

FedAvg
& 24.3 & 29.8 & 24.4 & 39.8 & \textbf{74.6} & 1052.0 & 34.2 \\

FedAdam
& 22.5 & 29.3 & 22.6 & 39.1 & 71.9 & 995.4 & 33.1 \\

FedProx
& 20.6 & 25.6 & 20.2 & 37.5 & 69.5 & 931.2 & 29.4 \\

MOON
& 26.1 & \underline{32.2} & 24.8 & 40.6 & 72.8 & 962.9 & 33.8 \\
\rowcolor{lightgray}
FedAvg-TA
& 26.4 & 28.5 & 22.8 & 40.2 & 71.0 & 949.6 & \textbf{35.2} \\
\rowcolor{lightgray}
FedAvg-TIEs
& \underline{28.7} & 32.1 & \underline{26.6} & 41.5 & \underline{74.2} & \textbf{1105.2} & \underline{34.3} \\
\rowcolor{lightgray}
FedAvg-DARE
& 26.8 & 29.2 & 19.3 & \underline{41.6} & 68.1 & 684.7 & 31.2 \\

\rowcolor{oursblue}
Fed-CMP
& \textbf{31.6} & \textbf{34.5} & \textbf{27.1} & \textbf{44.8} & 73.4 & \underline{1068.6} & \underline{34.3} \\
\bottomrule
\end{tabular}
\end{table*}

\begin{table*}[t]
\centering
\small
\setlength{\tabcolsep}{10pt}
\caption{Evaluation of MLLMs on multimodal benchmarks under the $\mathcal{M}_{Cross}$ setting.}\label{table_it_ti}
\begin{tabular}{lccccccc}
\toprule
Method & MM-Vet & MMBench & SEED & LLaVABench & POPE & MME & MMVP \\
\midrule
Local Training
& 27.8 & 24.9 & 25.1 & 36.7 & 62.1 & 999.5 & 31.2 \\

FedAvg
& 22.4 & 26.9 & 27.8 & 41.3 & \underline{75.6} & 1084.9 & \underline{34.8} \\

FedAdam
& 20.1 & 23.2 & 26.5 & 38.5 & 67.9 & 1033.6 & 33.1 \\

FedProx
& 19.3 & 22.6 & 20.7 & 39.8 & 71.2 & 1010.7 & 31.4 \\

MOON
& 23.7 & 27.4 & 28.2 & 42.1 & 74.1 & 956.3 & 33.8 \\
\rowcolor{lightgray}
FedAvg-TA
& 24.9 & 26.1 & 23.5 & 41.9 & 70.3 & 902.4 & \textbf{35.6} \\
\rowcolor{lightgray}
FedAvg-TIEs
& \underline{29.8} & \underline{31.9} & \underline{29.4} & 43.7 & \textbf{76.8} & \underline{1112.5} & 32.2 \\
\rowcolor{lightgray}
FedAvg-DARE
& 26.9 & 24.5 & 22.1 & \underline{46.3} & 68.7 & 836.8 & 27.2 \\

\rowcolor{oursblue}
Fed-CMP
& \textbf{33.4} & \textbf{34.8} & \textbf{31.6} & \textbf{48.2} & 75.1 & \textbf{1165.9} & 34.7 \\
\bottomrule
\end{tabular}
\end{table*}

\section{Additional Experimental Results}
\label{additional experiment}
In this section, we present additional experimental results under two clustering settings: Joint Image \& Text Clustering ($\mathcal{M}_{Joint}$) and Cross-Modal Image-Text Clustering ($\mathcal{M}_{Cross}$).
The experiment results are reported in Tables~\ref{table_it_it} and~\ref{table_it_ti}, respectively.

As shown in the tables, Fed-CMP consistently achieves the best performance on the majority of multimodal benchmarks.
Comparing these results with the $\mathcal{M}_{I-I}$ and $\mathcal{M}_{T-T}$ settings in the main paper, we observe a clear correlation between data heterogeneity and the performance advantage of our method.
Specifically, the degree of heterogeneity follows the order: $\mathcal{M}_{T-T} < \mathcal{M}_{cross} < \mathcal{M}_{I-I} \approx \mathcal{M}_{joint}$.
Under lower heterogeneity settings, performance gaps among methods are relatively narrow; as heterogeneity increases, the superiority of Fed-CMP becomes more pronounced.
This trend indicates that Fed-CMP is particularly effective in high-heterogeneity scenarios, providing a robust solution for enhancing model generalization in challenging non-IID environments.

\begin{algorithm}[t]
\caption{Fed-CMP}
\label{alg:fedmp}
\begin{algorithmic}[1]
\REQUIRE Number of clients $K$; Communication rounds $R$; Local datasets $\{\mathcal{D}_k\}_{k=1}^{K}$; Initial global projector $\mathbf{W}_0$; Momentum bounds $\beta_{\min}$, $\beta_{\max}$; Sensitivity $\lambda$; Alignment weight $\alpha$.
\ENSURE Final global projector $\mathbf{W}_R$

\STATE Server initializes and broadcasts $\mathbf{W}_0$ to all clients
\FOR{$r = 1, 2, \ldots, R$}
    \STATE \textcolor{gray}{\textit{Client Local Training}}
    \FOR{each client $k \in \{1, \ldots, K\}$ \textbf{in parallel}}
        \STATE Receive global projector $\mathbf{W}_{r-1}$ from server
        \STATE $\mathbf{W}_k \leftarrow \text{LocalTrain}(\mathbf{W}_{r-1}, \mathcal{D}_k)$
        \STATE Compute alignment error: $\mathrm{d}_k = \frac{1}{|\mathcal{D}_k|} \sum_{i \in \mathcal{D}_k} \left\| \tilde{\mathbf{x}}^{img}_{i,k} - \tilde{\mathbf{x}}^{t}_{i,k} \right\|_2$
        \STATE Send $(\mathbf{W}_k, \mathrm{d}_k)$ to server
    \ENDFOR
    
    \STATE \textcolor{gray}{\textit{Server: Canonical Reliability-Aware Aggregation (CRA)}}
    \STATE Concatenate: $\Delta \mathbf{W} = [\mathbf{W}_1, \mathbf{W}_2, \ldots, \mathbf{W}_K]$
    \STATE $\hat{\mathbf{U}}_r, \hat{\boldsymbol{\Sigma}}_r, [\mathbf{V}_1, \ldots, \mathbf{V}_K]^\top \leftarrow \mathrm{SVD}(\Delta \mathbf{W})$
    \FOR{each client $k \in \{1, \ldots, K\}$}
        \STATE $w_k = \frac{\left\| \mathbf{W}_k - \mathbf{W}_{r-1} \right\|^2 \cdot e^{(-\alpha \cdot \mathrm{d}_k)}}{\sum_{i=1}^{K} \left( \left\| \mathbf{W}_i - \mathbf{W}_{r-1} \right\|^2 \cdot e^{(-\alpha \cdot \mathrm{d}_i)} \right)}$
    \ENDFOR
    \STATE $\mathbf{V}_{\text{fused}} = \sum_{k=1}^{K} w_k \cdot \mathbf{V}_k$
    
    \STATE \textcolor{gray}{\textit{Server: Orthogonality-Preserved Momentum (OPM)}}
    \IF{$r = 1$}
        \STATE $\mathbf{U}_r \leftarrow \hat{\mathbf{U}}_r$; \quad $\boldsymbol{\Sigma}_r \leftarrow \hat{\boldsymbol{\Sigma}}_r$
    \ELSE
        \STATE $\beta \leftarrow \text{AdaptiveMomentum}(\hat{\mathbf{U}}_r, \mathbf{U}_{r-1}, \beta_{\min}, \beta_{\max})$
        \STATE $\mathbf{U}_r \leftarrow \text{OrthMomentum}(\hat{\mathbf{U}}_r, \mathbf{U}_{r-1}, \beta)$
        \STATE $\boldsymbol{\Sigma}_r \leftarrow \beta \cdot \boldsymbol{\Sigma}_{r-1} + (1 - \beta) \cdot \hat{\boldsymbol{\Sigma}}_r$
    \ENDIF
    
    \STATE $\mathbf{W}_r \leftarrow \mathbf{U}_r \boldsymbol{\Sigma}_r \mathbf{V}_{\text{fused}}^\top$
\ENDFOR
\STATE \textbf{return} $\mathbf{W}_R$
\end{algorithmic}
\end{algorithm}

\section{Algorithm Overview}
\label{sec:algorithm}
In this section, we summarize the complete Fed-CMP framework. 
Algorithm~\ref{alg:fedmp} presents the detailed procedure, which consists of three main phases per communication round: (1) client local training, (2) server Canonical Reliability-Aware Aggregation (CRA), and (3) server Orthogonality-Preserved Momentum (OPM) update.
During local training, each client performs autoregressive training on private image-text pairs and computes the local cross-modal alignment error.
Upon receiving local updates, the server first applies CRA to decouple client projectors into shared basis and client-specific coefficients, then fuses the client-specific coefficients via reliability-weighted fusion.
Subsequently, OPM applies momentum to the shared alignment basis, ensuring stable evolution across communication rounds while preserving the geometric structure of the semantic space.
Finally, the global projector is reconstructed and distributed to clients for the next round.

\section{Dataset Construction}\label{dataset}
To evaluate the efficacy of our method, we simulate a classical federated pretraining scenario involving multiple clients. We maximize the statistical heterogeneity by constructing a data partition where the pretraining data distribution varies significantly across different clients.

\subsection{Feture Extration}
We utilize the CC12M~\cite{cc12m} dataset as the source for training samples. Our pipeline involves: 1) extracting multimodal features using a pre-trained CLIP-like encoder~\cite{clip} and 2) subsequently partitioning the data based on semantic clustering.

Formally, let the image-text dataset be denoted as $\mathcal{D} = \{ (I_i, T_i) \}_{i=1}^{N}$, where $I_i$ represents the image and $T_i$ represents the corresponding text caption for the $i$-th sample. We employ SigLIP2-L-512~\cite{siglip2} as our image and text feature extractors $\mathcal{E} = \{\mathcal{E}_{\text{img}}, \mathcal{E}_{\text{txt}}\}$ to obtain high-dimensional embeddings for both modalities.
For a given sample $x_i = (I_i, T_i)$, the feature extraction process is defined as:
\begin{equation}
    \mathbf{v}_I^{(i)} = \mathcal{E}_{\text{img}}(I_i), \quad \mathbf{v}_T^{(i)} = \mathcal{E}_{\text{txt}}(T_i)
\end{equation}
where $\mathbf{v}_I^{(i)}, \mathbf{v}_T^{(i)} \in \mathbb{R}^d$ are the normalized image and text embeddings, respectively.

\subsection{Heterogeneous Clustering}
To simulate non-IID distributions, we partition the dataset into $K=5$ distinct clusters. We employ a clustering algorithm that groups samples based on cosine similarity, ensuring alignment with the SigLIP's contrastive embedding space. We propose four distinct metrics to define the similarity measure $\mathcal{S}(x_i, x_j)$ used for clustering:

\begin{enumerate}
    \item \textit{Image-Image Clustering ($\mathcal{M}_{I\text{-}I}$):} Clustering is performed solely based on visual semantic similarity.
    \begin{equation}
        \mathcal{S}(x_i, x_j) = \mathbf{v}_I^{(i)} \cdot \mathbf{v}_I^{(j)}
        \label{eqn:image-image}
    \end{equation}

    \item \textit{Text-Text Clustering ($\mathcal{M}_{T\text{-}T}$):} Clustering relies only on the textual captions, grouping data based on linguistic semantics.
    \begin{equation}
        \mathcal{S}(x_i, x_j) = \mathbf{v}_T^{(i)} \cdot \mathbf{v}_T^{(j)}
        \label{eqn:text-text}
    \end{equation}

    \item \textit{Joint Image \& Text Clustering ($\mathcal{M}_{Joint}$):} This metric considers intra-modality similarities for both modalities simultaneously.
    \begin{equation}
        \mathcal{S}(x_i, x_j) = \mathbf{v}_I^{(i)} \cdot \mathbf{v}_I^{(j)} + \mathbf{v}_T^{(i)} \cdot \mathbf{v}_T^{(j)}
        \label{eqn:joint}
    \end{equation}

    \item \textit{Cross-Modal Image-Text Clustering ($\mathcal{M}_{Cross}$):} To leverage the alignment capabilities of SigLIP, this metric clusters based on cross-modal similarity.
    \begin{equation}
        \mathcal{S}(x_i, x_j) = \mathbf{v}_I^{(i)} \cdot \mathbf{v}_T^{(j)}
        \label{eqn:cross}
    \end{equation}
\end{enumerate}

\begin{table}[t]
    \centering
    \caption{Distribution Statistics for pretraining data clustering.}
    \label{tab:clustering_distribution}
    \setlength{\tabcolsep}{2pt}
    \begin{tabular}{l|c|ccccc}
        \toprule
        \textbf{Method} & \textbf{Metric} & \textbf{Cluster 0} & \textbf{Cluster 1} & \textbf{Cluster 2} & \textbf{Cluster 3} & \textbf{Cluster 4} \\
        \midrule
        
        \multirow{2}{*}{\textbf{Image-Image}} 
        & Count & 1M & 1M & 1M & 1M & 1M \\
        & Percentage & 20.00\% & 20.00\% & 20.00\% & 20.00\% & 20.00\% \\
        \midrule
        
        \multirow{2}{*}{\textbf{Text-Text}}   
        & Count & 368K   & 1M & 556K   & 1M & 680K \\
        & Percentage & 10.22\% & 27.73\% & 15.44\% & 27.73\% & 18.88\% \\
        \midrule
        
        \multirow{2}{*}{\textbf{Joint-Modal}} 
        & Count & 1M & 615K   & 1M & 1M & 979K \\
        & Percentage & 21.77\% & 13.39\% & 21.77\% & 21.77\% & 21.31\% \\
        \midrule
        
        \multirow{2}{*}{\textbf{Cross-Modal}} 
        & Count & 1M & 307K   & 1M & 1M & 663K \\
        & Percentage & 25.18\% & 7.74\%  & 25.18\% & 25.18\% & 16.71\% \\
        
        \bottomrule
    \end{tabular}
\end{table}

Following the clustering process, we prune samples distant from the cluster centroids to cap each cluster at a maximum of 1 million samples. This step sharpens the distributional boundaries between clusters, further enlarge the statistical heterogeneity among clients. To this end, we formulate 4 distinct datasets, where each consists of 5 clusters, with approximately 1 million samples per cluster. Table~\ref{tab:clustering_distribution} presents the detailed statistics for each dataset.

\begin{figure*}[t]
\centering
    \includegraphics[width=0.95\linewidth]{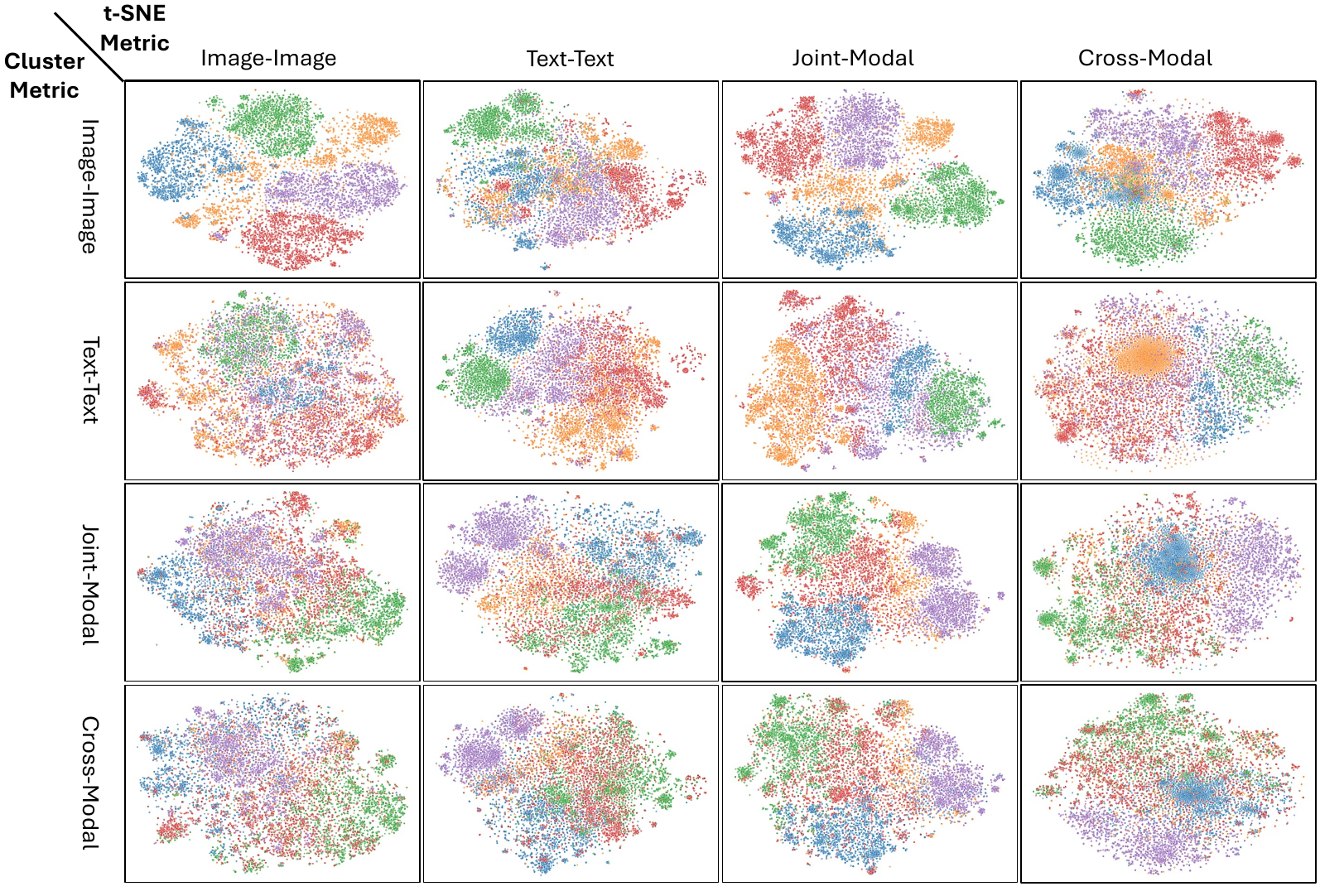}
    \caption{t-SNE visualization of clustering results using four distinct similarity metrics. 
}\label{fig:tsne_full}
\end{figure*}




\subsection{Analysis}
We visualize the four sub-datasets using t-SNE~\cite{tsne}, projecting the data distributions based on the distance metrics defined in Eqns.~\eqref{eqn:image-image}--\eqref{eqn:cross}. The results are presented in Figure~\ref{fig:tsne_full}. Based on these visualizations, we observe three key phenomena:

\begin{enumerate}
    \item \textbf{Metric Consistency:} As expected, each dataset exhibits the most distinct cluster separation when visualized using the specific metric employed for its construction. This confirms that our clustering approach effectively captures the intended semantic structures.
    
    \item \textbf{Visual Discriminability:} The clusters formed via Image-Image clustering ($\mathcal{M}_{I\text{-}I}$) display the sharpest boundaries. This suggests that the visual embedding space of SigLIP possesses a more granular and discriminative manifold structure compared to the textual space, allowing for clearer separation of semantic concepts.
    
    \item \textbf{Cross-Modal Ambiguity:} The separation in the cross-modal setting ($\mathcal{M}_{Cross}$) is less distinct compared to single-modal clustering. This is likely because Eq.~\eqref{eqn:cross} measures the \textit{alignment intensity} between image and text rather than their inherent semantic content. Consequently, samples may possess high alignment scores yet share underlying semantic features with neighboring clusters, leading to softer boundaries in the projected space.
\end{enumerate}

\begin{figure}[t]
\centering
    \includegraphics[width=0.95\linewidth]{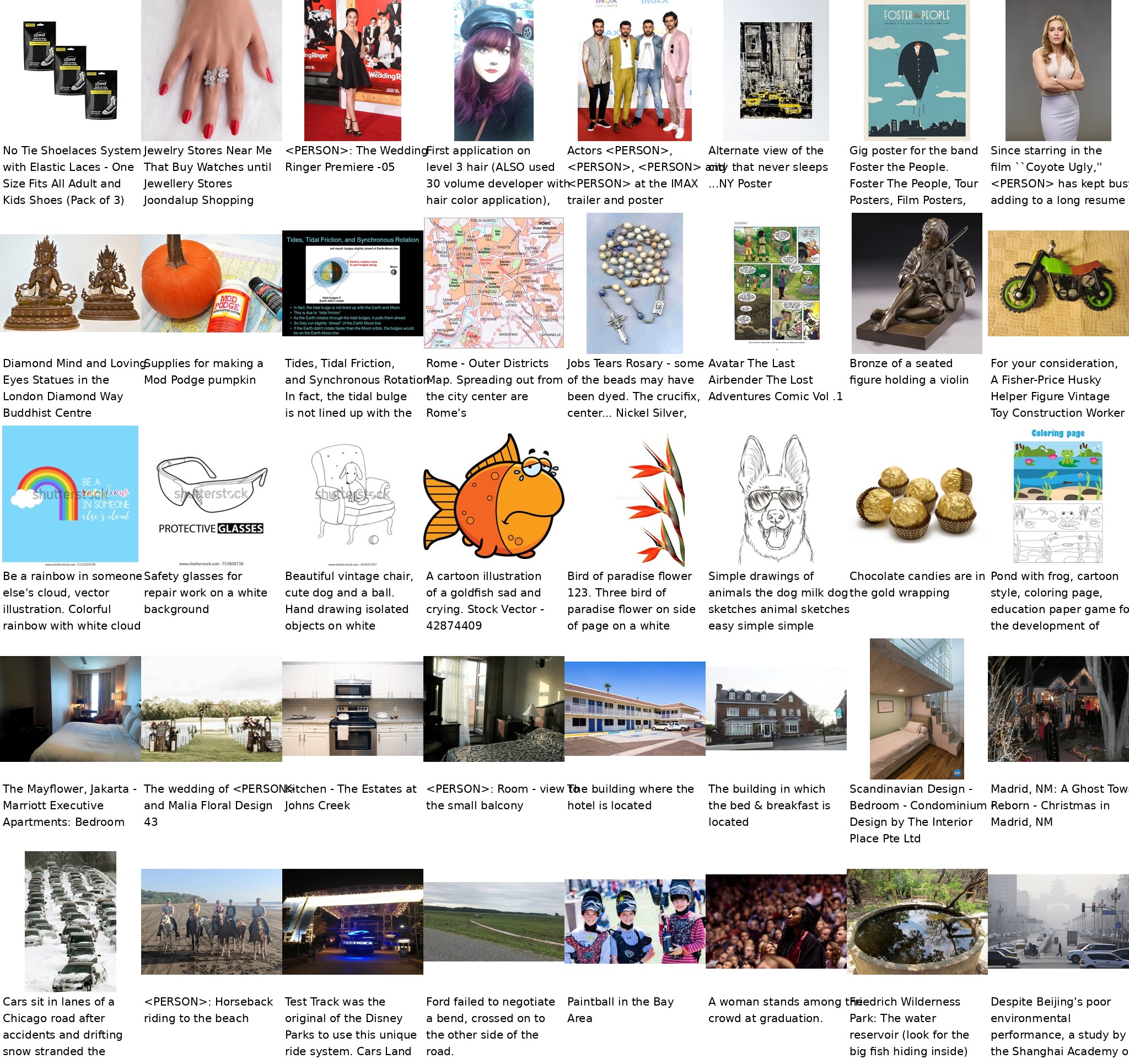}
    \caption{Randomly sampled image-text pairs from the image-image clustering results. Each row represents a single cluster. 
}\label{fig:image-image}
\end{figure}

\begin{figure}[t]
\centering
    \includegraphics[width=0.95\linewidth]{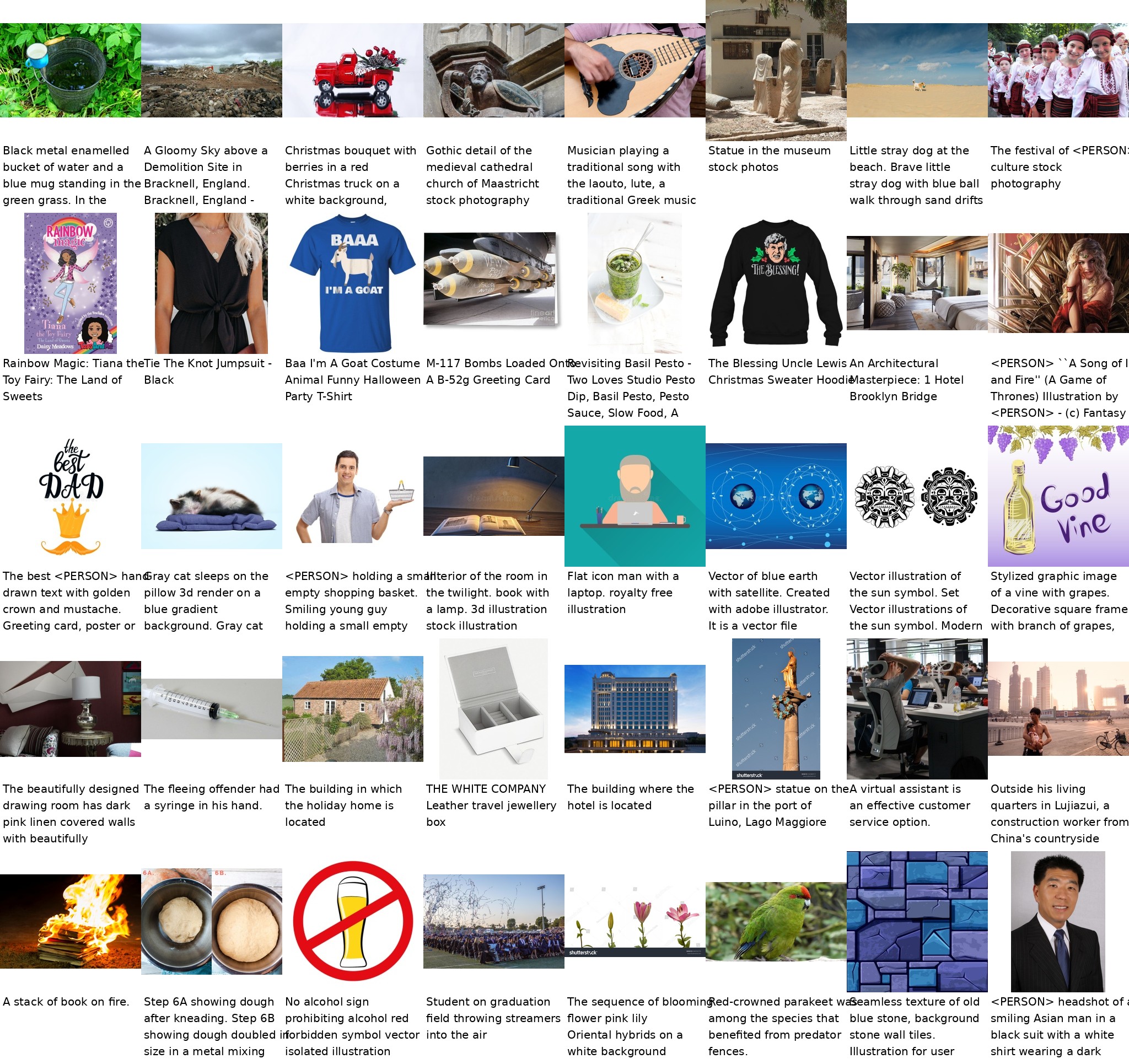}
    \caption{Randomly sampled image-text pairs from the text-text clustering results. Each row represents a single cluster.
}\label{fig:text-text}
\end{figure}

\begin{figure}[t]
\centering
    \includegraphics[width=0.95\linewidth]{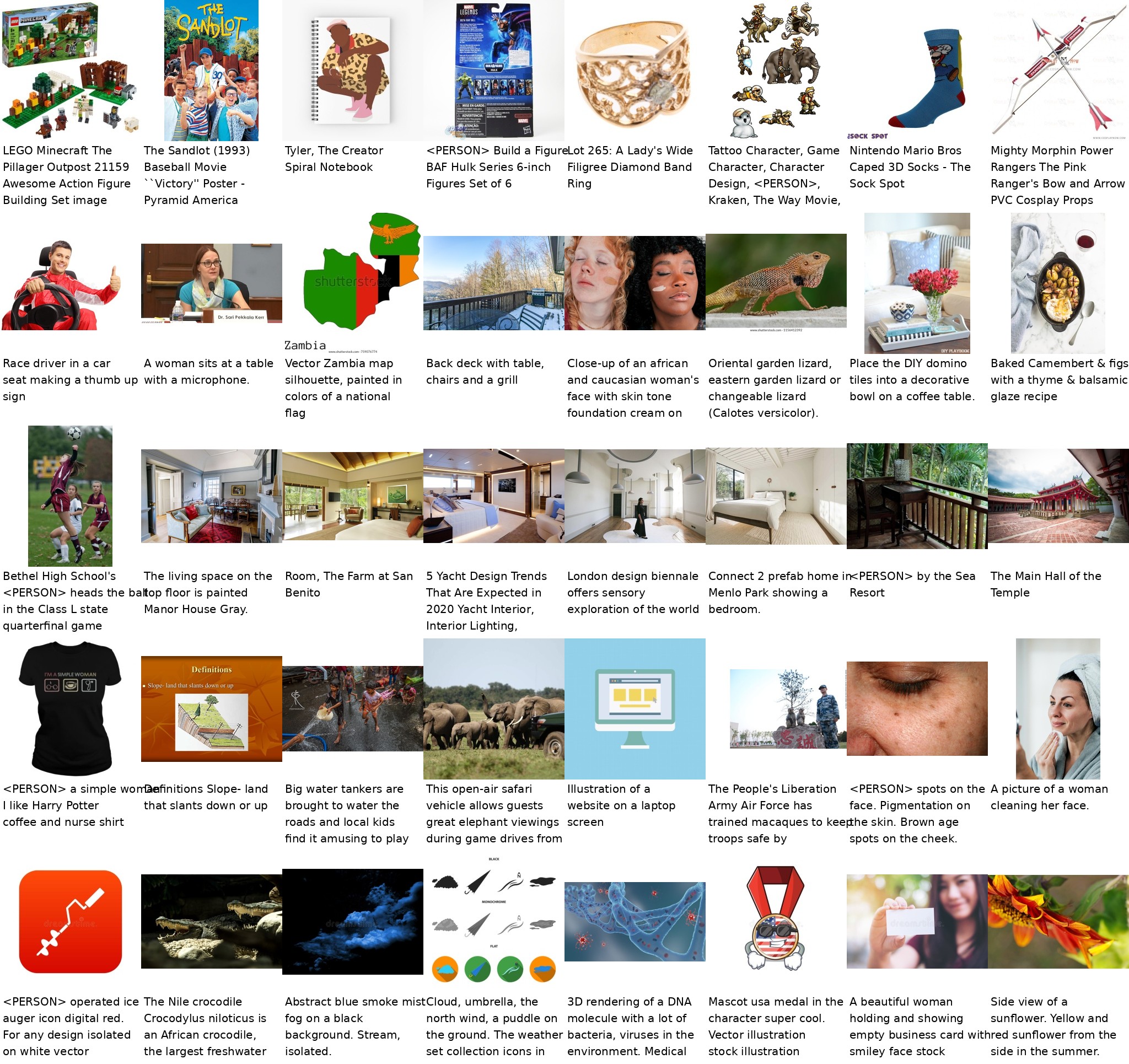}
    \caption{Randomly sampled image-text pairs from the joint image and text clustering results. Each row represents a single cluster.
}\label{fig:joint}
\end{figure}

\begin{figure}[t]
\centering
    \includegraphics[width=0.95\linewidth]{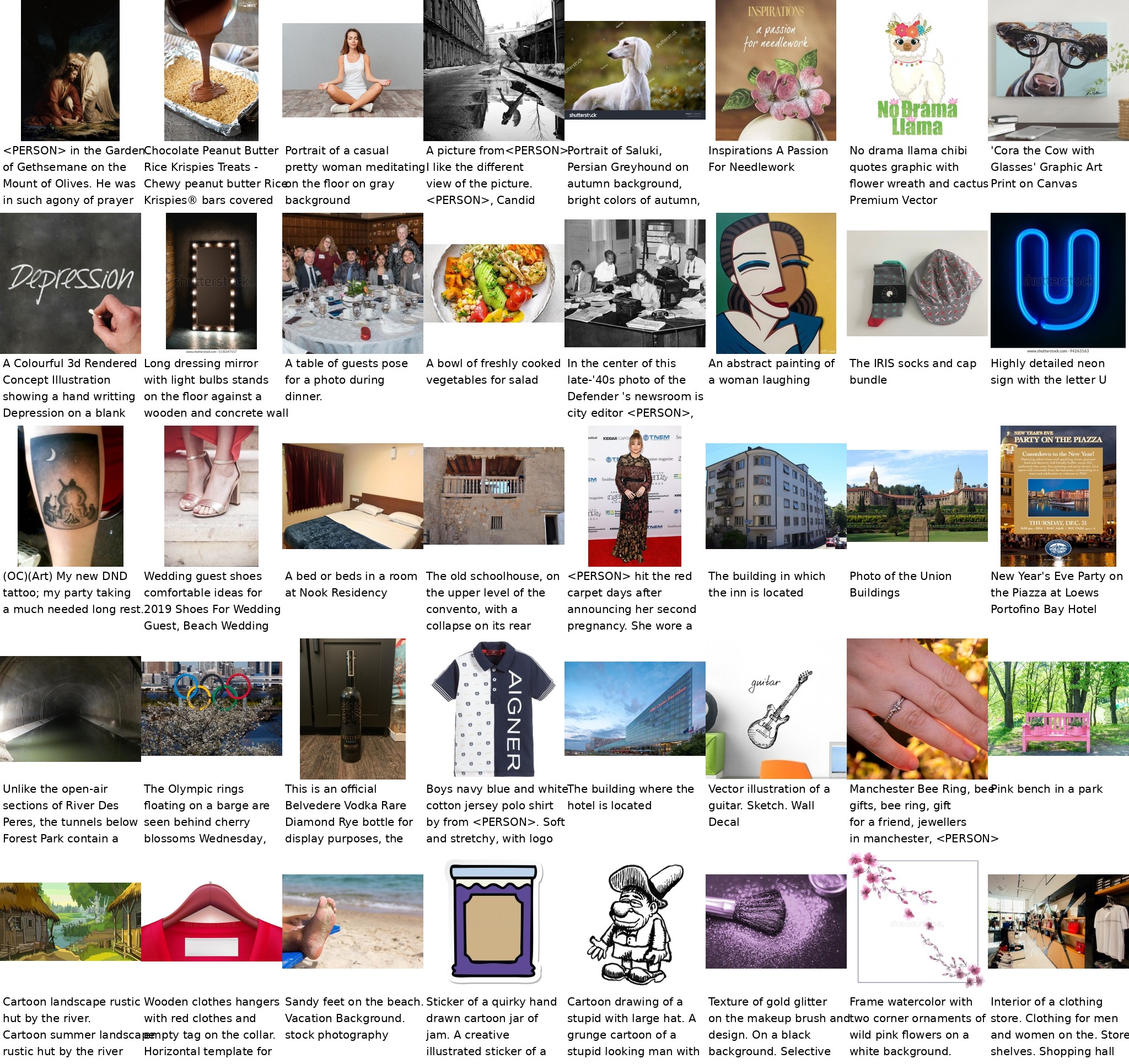}
    \caption{Randomly sampled image-text pairs from the cross-modal image-text clustering results. Each row represents a single cluster.
}\label{fig:cross}
\end{figure}

Furthermore, we inspect the qualitative content of these clusters. Figure~\ref{fig:image-image} illustrates samples from the dataset constructed via $\mathcal{M}_{I\text{-}I}$. The five clusters demonstrate clear thematic divergence, predominantly capturing concepts related to: 1) fashion and clothing, 2) text-rich imagery, 3) sketches and drawings, 4) household and indoor scenes, and 5) crowded environments. Similar semantic consistency is observed for the other metrics in Figures~\ref{fig:text-text}, \ref{fig:joint}, and \ref{fig:cross}.

\section{Limitations}
\label{sec:limitations}
While Fed-CMP demonstrates promising results in federated MLLM pre-training, several limitations remain.
First, our framework focuses only on the cross-modal projector, leaving the visual encoder and LLM unchanged.
How to perform full or partial fine-tuning of larger components in a federated environment remains an open challenge.
Second, the SVD-based decomposition in CRA introduces additional computational overhead on the server side, potentially becoming a bottleneck when scaling to a large number of clients.
Third, our experiments were conducted in a simulated federated environment with controlled data partitioning.
Further investigation is needed to address real-world deployment scenarios involving heterogeneous devices, asynchronous communication, and client downtime.

\end{document}